\documentclass{article}

\usepackage{PRIMEarxiv}

\usepackage[ruled,vlined,linesnumbered,noend]{algorithm2e}
\usepackage{amsmath}
\usepackage{booktabs}

\DeclareMathOperator*{\argmin}{arg\,min}
\DeclareMathOperator*{\argmax}{arg\,max}

\DeclareMathOperator{\Min}{min}

\usepackage[utf8]{inputenc} 
\usepackage[T1]{fontenc}    
\usepackage{hyperref}       
\usepackage{url}            
\usepackage{booktabs}       
\usepackage{amsfonts}       
\usepackage{nicefrac}       
\usepackage{microtype}      
\usepackage{lipsum}
\usepackage{fancyhdr}       
\usepackage{graphicx}       
\graphicspath{{media/}}     

\title{AD-NEv++ : The multi-architecture neuroevolution-based multivariate anomaly detection framework
\thanks{\textbf{Paper accepted at GECCO 2024}} 
}

\author{
  Marcin Pietroń \\
  Institute of Electronics\\
  AGH\\
  Kraków\\
  \texttt{pietron@agh.edu.pl} \\
  \And
  Dominik Żurek \\
  Department of Computer Science \\
  AGH \\
  Kraków \\
  \texttt{dzurek@agh.edu.pl} \\
  \And
  Kamil Faber \\
  Department of Computer Science \\
  AGH \\
  Kraków \\
  \texttt{kfaber@agh.edu.pl} \\
   \And
  Roberto Corizzo \\
  Department of Computer Science \\
  American University \\
  Washington D.C.\\
  \texttt{rcorizzo@american.edu} \\
}

\begin{document}
\maketitle

\begin{abstract}
Anomaly detection tools and methods enable key analytical capabilities in modern cyberphysical and sensor-based systems. Despite the fast-paced development in deep learning architectures for anomaly detection, model optimization for a given dataset 
is a cumbersome and time-consuming process.
Neuroevolution could be an effective and efficient solution to this problem, as a fully automated search method for learning optimal neural networks, 
supporting both gradient and non-gradient fine tuning.
However, existing frameworks incorporating neuroevolution
lack of support for new layers and architectures and are typically limited to convolutional and LSTM layers. 
In this paper we propose AD-NEv++, a three-stage neuroevolution-based method that synergically combines subspace evolution, model evolution, and fine-tuning. Our method overcomes the limitations of existing approaches by optimizing the mutation operator in the neuroevolution process, while supporting a wide spectrum of neural layers, including attention, dense, and graph convolutional layers.
Our extensive experimental evaluation was conducted with 
widely adopted multivariate anomaly detection benchmark datasets, and showed that the models generated by AD-NEv++ outperform well-known deep learning architectures and neuroevolution-based approaches for anomaly detection. Moreover, results show that AD-NEv++ can improve and outperform the state-of-the-art GNN (Graph Neural Networks) model architecture in all anomaly detection benchmarks.
\end{abstract}


\section{Introduction}
Anomaly detection tools are of paramount importance in modern cyberphysical and sensor-based systems that involve generating high-dimensional time-series data. 
These systems must be continuously monitored to perform maintenance in a timely manner and prevent expensive failures. 
In anomaly detection, a common scenario is to have abundant availability of normal data acquired during monitoring, and a scarcity of labeled anomalies. 
To this end, the semi-supervised anomaly detection learning setting is the most applicable in real-world applications, where model training is conducted exclusively, resorting to normal data \cite{pang2021}. Alternatively, unsupervised learning can be adopted where training data is mostly normal, but possibly contains a small number of  anomalies. 
Considering recent anomaly detection works in such learning settings, deep learning methods and, specifically, autoencoder-based approaches, have shown to be the particularly robust \cite{malhotra2016lstmbased}\cite{zhou}\cite{zhang2019}\cite{su2019}\cite{USAD}\cite{tariq2019}\cite{tuor2017}\cite{zheng}. 
Among these, graph-based autoencoders were identified as the best-performing approach in well-known benchmark datasets \cite{graph-nn} \cite{ahmed2021graph} \cite{feng2022full} \cite{ren2024graph}.
Although deep auto-encoder models are a viable option for this task, one crucial limitation is the difficulty in identifying a proper configuration, including the definition of the model architecture, and an extensive hyperparameter tuning required to achieve high performance.
To this end, neuroevolution can ease the burden of this cumbersome task. In the literature, classical approaches include NEAT \cite{NEAT}, HyperNEAT \cite{stanley} and coDeepNEAT \cite{neuro_evolving}, which aim to optimize parameters, model architectures, or both.
The first attempt at neuroevolution-based anomaly detection was done in \cite{ADNEV}, which considers the joint optimization of model architectures with feature subspaces and model weights. 
However, the optimization is carried out for specific layers, such as convolutional (CNN) and Long-Short Term Memory (LSTM), without supporting recent graph-based auto-encoder architectures. Another limitation is the lack of optional layers such as attention, skip connection, and dense connections, which have been shown to enhance performance in several downstream tasks \cite{zhang2023mdu,agarwal2023attention,yan2023memory}. 
In this paper, our main contribution is to extend the ADNEv \cite{ADNEV} by proposing ADNEv++. The proposed method significantly extends ADNEv's capabilities by
incorporating graph autoencoders as a new type of neural architecture in the neuroevolution process, and by 
defining a new layer of abstraction to consider optional layers, such as attention, skip connection, and dense connections.

\begin{figure*}[h!]
  \centering
  \includegraphics[width=0.995\linewidth,clip=true]{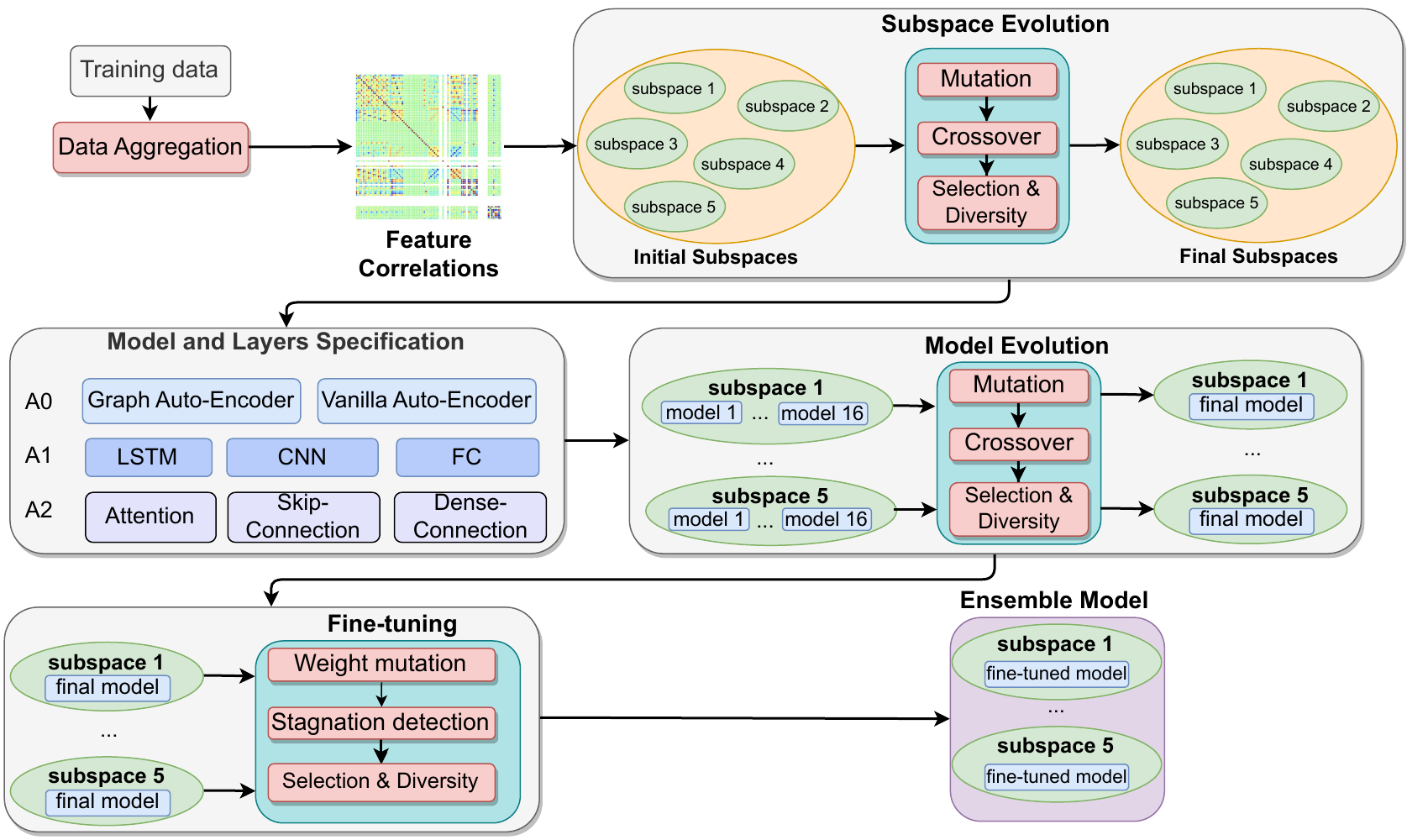}
    \caption{AD-NEv++ framework architecture. Time series data is used to train and evaluate a set of models during neuroevolution through different levels. The framework returns an optimized ensemble model based on an initially specified model architecture supporting choices at different levels of abstraction: basic model (A0), specific layers (A1), and optional layers (A2). The framework extends and generalizes AD-NEv's optimization workflow with support for a larger class of models,  including graph auto-encoders and layers such as Attention, Skip-Connection, and Dense-Connection.}
    \label{fig:neral_schema}
\end{figure*}

Our approach involves the simultaneous evolution of two populations: models and subspaces. The former consists of neural network architectures that evolve during the neuroevolution process, whereas the latter define subsets of input features. 
After the neuroevolution process, the framework yields a bagging technique-based ensemble model derived from single optimized architectures.
Our experiments show that the multi-level neuroevolution approach of ADNEv++ allows us to identify robust and optimized autoencoder architectures for anomaly detection.
The results show that the consideration of graph-based architecture and the abstraction to support optional layers in ADNEv++ enhance the flexibility of the framework and allow us to significantly and consistently outperform ADNEv, as well as other popular baselines, on all benchmark datasets considered in our analysis.

\section{Related works}

In this section, we delve into a detailed analysis of anomaly detection methods tailored to multivariate time-series data, alongside an exploration of neuroevolution methods. This analysis is grounded in the context of recent advancements and comprehensive reviews in the field of anomaly detection. The significant surveys include a broad overview of general anomaly detection techniques \cite{chandola2009}, analysis of deep learning-based approaches to anomaly detection \cite{pang2021}, \cite{chalapathy2021}, \cite{choi2021}, \cite{landauer2023}, \cite{fernando2021}, and analysis of unsupervised time-series anomaly detection \cite{blaz2020}. These surveys collectively present a comprehensive view of the current state and emerging trends in deep learning applications for anomaly detection.

Significant trends include use of autoencoder-based methods, which become prominent in a number of real world applications, such as cybersecurity \cite{torabi2023}\cite{faber2023vlad} \cite{faber2023distributed}, energy \cite{corizzo2021spatially}, physics \cite{finke2021}, and medical imaging \cite{shvetsova2021}. Two exemplar methods are USAD \cite{USAD} and UAE \cite{garg}. The USAD model extends beyond the standard autoencoder by incorporating an additional discriminator and loss extension.
In their comparative analysis of multivariate anomaly detection models, the authors of \cite{garg} presents the specifics of the UAE model. The UAE consists of multiple autoencoders, with each encoder connected to a distinct feature. These encoders are structured as multi-layer perceptrons, with the number of nodes in each layer corresponding to the number of time steps (window size). The model achieves dimensionality reduction in the latent space by a factor of two. The decoder, designed as a mirror image of the encoder, utilizes a $tanh$ activation function.

Other types of approaches include: i) LSTM-based methods, such as NASA-LSTM \cite{nasa-lstm}, LSTM-AE \cite{garg} (based on \cite{malhotra2016lstmbased}), and LSTM-VAE \cite{Chen2019SequentialVF}; ii) CNN-based methods, such as temporal convolutional AE (TCN AE) \cite{bai2018empirical}; iii) GAN-based methods, such as OCAN \cite{zheng}, and BeatGAN \cite{zhou}; iv) graph neural network-based approaches, such as \cite{graph-nn}. Moreover, there are some hybrid methods, such as MSCRED \cite{mscred}, DAGMM \cite{zong2018}, and OmniAnomaly \cite{su}.

Among auto-encoder architectures, graph-based approaches have shown state-of-the-art performance in anomaly detection. 
The approach in \cite{graph-nn} presents a dense graph neural network where each node represents a single feature, and edges allow to represent data exchanged between different nodes. 
The graph-based modeling capabilities yielded a significant performance boost over other state-of-the-art methods with very well-known multivariate time series datasets.
A mirror temporal graph autoencoder framework for traffic anomaly detection is proposed in \cite{ren2024graph}, where a temporal convolutional module and a graph convolutional gate recurrent unit cell with Gaussian kernel functions are used to capture hidden node-to-node features in the traffic network.
Authors in \cite{feng2022full} proposed a full graph autoencoder leveraging to perform semi-supervised anomaly detection in large-scale IIoT systems. To this end, the authors extend the classical graph layers with a variational model to fully learn the graph representation.
A different approach is proposed in \cite{ahmed2021graph}, where a minimum spanning tree (MST) and a graph-based algorithm are proposed to approximate the local neighborhood structure, giving place to a graph-regularized autoencoder. 

From another perspective, approaches involving attention, skip-connection, and dense-connection layers have shown to be promising to enhance model performance in a variety of downstream tasks.
Authors in \cite{agarwal2023attention} adopt attention-based fusion of encoder and decoder features to address monocular depth prediction,  and show that this approach allows to improve the learning embedding and provides an improved model generalization performance.
A skip-connected deep autoencoder for anomaly detection is proposed in \cite{yan2023memory} to deal with multi-source data fusion of  rocket engines, where each layer of the encoder and decoder has a skip connection to fully extract multi-scale features of the normal sample in a multi-dimensional space.
The work in \cite{zhang2023mdu} proposes a multi-scale densely connected U-net for biomedical image segmentation, where dense-connections are used to directly fuse the neighboring different scale feature maps from both higher layers and lower layers, strengthening feature propagation at a given layer, and improving the information flow between the encoder and decoder counterparts of the model.



An existing drawback of these anomaly detection methods, as well as deep learning methods involving attention, skip-connections, and dense connections, is that they present a cumbersome setup process. Since they do not perform automatic model optimization, they require a significant manual effort to identify the right architecture and tune it for the right domain and dataset. 
This limitation may be tackled with neuroevolution approaches, which can support model performance in downstream  machine learning tasks by improving model architectures \cite{ADNEV} \cite{neuroevol_overview}, \cite{danilo2016}, \cite{ming2018} and \cite{huang2022}. 
An examples of this type of approach is \cite{neuro_evolving}, where a two-level neuroevolution strategy based on the co-deep NEAT algorithm is used to outperform human-designed models in specific tasks such as  image classification and  language modeling. 
In \cite{huang2022}, a novel deep reinforcement learning-based framework optimized by neuroevolution is proposed for electrocardiogram time-series signal. 
Self-organizing map-based neuroevolution for the carpool service problem is proposed in \cite{ming2018}. 

The authors in \cite{dimanov2021} propose a neuroevolution method to optimize the architecture and hyperparameters of convolutional autoencoders.  
Results on image datasets show that data can be compressed by a factor of 10 or higher, while still retaining salient information. 
Another example in the context of autoencoder models is shown in \cite{okada2017}, where a genetic algorithm is adopted to evolve their architecture. The authors found that reconstructions generated by these models are more accurate than manually curated autoencoders with more hidden units. 
A first attempt at developing a co-evolutionary neuroevolution-based multivariate anomaly detection system is presented in \cite{faber2021}. 
One critical drawback of this approach is, however, that the optimization of subspaces and models takes place separately. This design choice limits the capability of the neuroevolution process, forcing the model to compromise in order to handle all subspaces simultaneously and resulting in a potential reduction of its anomaly detection accuracy. Another limitation is the lack of fine-tuning capabilities. 
Fine-tuning is a popular technique to improve the accuracy of the pre-trained models. The most popular technique is gradient-based fine-tuning \cite{zhou2021}, \cite{ro2021}. However, a non-gradient approach can also yield significant improvements \cite{nagel2020}. 
The AD-NEv method \cite{ADNEV}  is the first framework for neuroevolution-based anomaly detection that incorporates the joint optimization of subspaces and models, and performs non-gradient fine-tuning. However, its capabilities are limited to auto-encoders with CNN and LSTM layers. To this end, this paper proposes AD-NEv++, which significantly extends AD-NEv by providing added support for graph-based autoencoder architectures and optional layers, such as attention, skip-connection, and dense-connection. In the remainder of the paper, we demonstrate that the proposed methodology works effectively with different model architectures and can effectively use a wide spectrum of layers that significantly outperform AD-NEv, as well as other popular baselines.

\section{Method}

In this section, we describe our AD-NEv++ method in detail. The method is graphically depicted in Figure \ref{fig:neral_schema}. We highlight that the AD-NEv++ framework presents the following novel characteristics in the neuroevolution process:

\begin{itemize}
    \item{incorporating new types of layers, originally not supported in the AD-NEv framework, such as multi-head attention and dense layers;}
    \item{updating the mutation operator to provide possibility of adding skip-connections and dense-connections between encoder and decoder layers (see Figure \ref{fig:dense});}
    \item{Adding graph neural autoencoders as a new supports type of neural model architecture in neuroevolution process.}
\end{itemize}

All added characteristics are provided in the second stage of the neuroevolution process (see Figure \ref{fig:neral_schema}). At the highest level of abstraction (A0), the basic architecture is selected in the model evolution process: vanilla autoencoder, or graph autoencoder. The next level (A1) is the selection of the type of autoencoder layers (LSTM, CNN, FC). The last level (A2) allows for the addition of optional layers in the form of skip and dense connections, as well as an attention layer to the encoder part.

\subsection{Autoencoders}
AD-NEv++ leverages a deep neural network architecture consisting of vanilla and graph-based autoencoders as base models for the neuroevolution process. 
The vanilla autoencoder consists of two parts: an encoder \textit{E} and a decoder \textit{D}. The encoder learns how to efficiently compress and encode the input data \textit{X} in a new representation with reduced dimensionality -- latent variables \textit{Z}. The decoder learns how to reconstruct the latent variables \textit{Z}. 

\begin{equation}
    AE(X) = D(Z),   Z = E(X).
\end{equation}

\begin{figure}[h!]
  \centering
  \includegraphics[width=0.65\linewidth,clip=true]{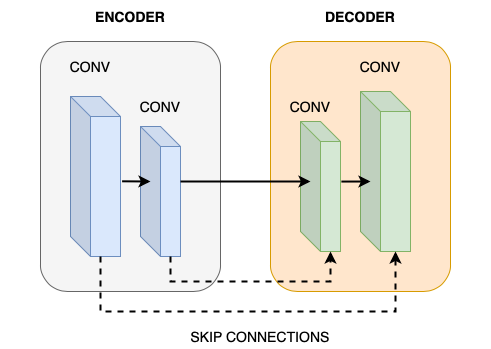}
    \caption{Addition of skip connections to a vanilla auto-encoder model with convolutional (CNN) layers.}
    \label{fig:dense}
\end{figure}

The vanilla autoencoder (A0) in AD-NEv++ can consist of fully connected (FC), convolutional (CNN), and recurrent (LSTM) layers (A1). As an option the attention mechanism can be added between encoder layers. Additionally, skip and dense connections can be applied between encoder and corresponding (mirror) decoder layers (A2) as shown in  Figure \ref{fig:dense}. In case of skip connections this process can be formalized as:
\begin{equation}
    y_{D_{i}} = D_{i}(E_{-i}(x) + D_{i-1}(x)).
\end{equation}

the $y_{D_{i}}$ is the output of the i-th decoder layer. The input to this layer is formed as a sum of the output of corresponding encoder layer and the output of previous decoder layer.

In case of dense connections the formalism is as follows:

\begin{equation}
    y_{D_{i}} = D_{i}([E_{-i}(x), D_{i-1}(x)]).
\end{equation}

The input of the i-th decoder is the concatenation of the output of previous decoder layer and output of corresponding encoder layer.

A distinctive feature of AD-NEv++ is the added support for Graph Neural Network autoencoders. These models consist of nodes and edges, where each node represents the single input sensor ($x_{i}$ - single sensor signal, $S$ - set of input sensors). Each node $j = 1, 2, \dots, N$ is processed by the $g_{j}$ layer:
\begin{equation}
    G(x) = \{g_{0}(x),..., g_{N}(x)\}.
\end{equation}

The $g_{j}$-th node processing layer consists of $f_{j}$ which processes the $j$ sensor, and $m_k$ which processes its neighbours. Finally, it returns the reconstructed input sensor: 
\begin{equation}
    g_{j}(x) = f_{j}(x_{j}) + \sum_{k}^{S} m_{k}(x_{k}).
\end{equation}

All models are trained to minimize the \textit{reconstruction loss}, which corresponds to minimizing the difference between the decoder's output and the original input data. It can be expressed as:
\begin{equation}
    \mathcal{L}(X, \hat{X}) = ||X - AE(X)||_2.
    \label{eq:autoencoder}
\end{equation}
When dealing with time series data, the error is computed at each time point (norm between input and multivariate reconstructed vector), as in  Equation \ref{eq:error}:
\begin{equation}
    Er_t = ||X_t - AE(x_t)||_2.
    \label{eq:error}
\end{equation}

\subsection{Model evolution}
Compared to AD-NEv, the most significant changes in the proposed framework can be identified in the second stage of model evolution. AD-NEv++ presents added support for new types of layers, new connections between layers, and a completely new architecture in addition to vanilla auto-encoders, i.e. the graph neural network model. In Algorithm \ref{alg:model_mutate}, the new characteristics introduced in AD-NEv++ can be identified at Lines 21-29. On Lines 21-23, the process adds a skip connection to the existing model. The $D'[-l]_{in}$ is an input of $D'[-l]$ decoder layer. The input is a sum of the outputs of the corresponding encoder layer ($F[l](x)$) and previous decoder layer ($D[-l-1](x)$). Furthermore, in Lines 24-26, a dense connection is added. The number of input channels of $D'[-l]$ layer is equal to the sum output channels of previous decoder layer $D[-l-1]$ and corresponding encoder layer $F[l]$. Finally, in Lines 27-29, we showcase the insertion of an attention layer at a specific position of the model. In case of the graph-based autoencoder the input and output dimensions of graph layers can be mutated. The window size is the parameter which can be modified in both autoencoder architectures.  
In order to extend the support of the crossover operation to a wider spectrum of model architectures, we performed one required modification. To this end, the crossover is placed as the second condition (see Algorithm \ref{alg:model_cross}, Line 9). Changing the length of models only occurs for a vanilla autoencoder. For graph autoencoders, instead, the number of layers is not exchanged, as we empirically observed that it did not provide any substantial benefit to the model's performance.



\begin{algorithm}
\KwIn{F - a model}
\KwIn{$w_{max}$ - maximum window size, $L_{max}$ - maximum number of conv, lstm, fully connected layers}
$F' \gets copy(F)$\;
sample $m$ from  $\{0,1,2,3,4,5\}$ \tcp*{mutation types}
\If{$m = 0$} { \tcp{mutate the number of channels in a layer} 
    sample $l$ from $\{0, \dots, L_{F}-1 \}$\;
    $c'$ = $randint(F[l]_{ic}, F[l+1]_{ic})$\;
    $F'[l]_{oc}$ $\gets$ $c'$\;
    $F'[l+1]_{ic}$ $\gets$ $c'$\;
}

\If{$m = 1$ and F is vanilla AE} { \tcp{reduce the length of the model}
    sample $l$ from $\{0, \dots, L_{max} \}$\;
    \uIf{$l < L_F$} {
    \For{$k \in [l+1,\dots,L_{F}]$} {
        $F'[k] \gets \emptyset$ \;
    }
    }
    \uElse{\For{$k \in [L_F,\dots,l-1]$} {
        $c'$ = $randint(F[k]_{oc}, F[k]_{oc}+c_{max})$\;
        $F'[k+1]_{ic}$ $\gets$ $F[k]_{oc}$\;
        $F'[k+1]_{oc}$ $\gets$ $c'$\;
    }}
    
}
\If{$m = 2$} { \tcp{mutate window size}
    sample $w$ from $\{1,\dots,w_{max}\}$\;
    $F'[0]_{ic}$ $\gets$ $w$\;
}

\If{$m = 3$ and F is vanilla AE} { \tcp{add skip connection}
    sample $l$ from $\{0, \dots, L_{max} \}$\;
    $D'[-l]_{in}$ = $F[l](x)$ + $D[-l-1](x)$\;
}
\If{$m = 4$ and F is vanilla AE} { \tcp{add dense connection}
    sample $l$ from $\{0, \dots, L_{max} \}$\;
    $D'[-l]_{ic}$ = $F[l]_{oc} + D[-l-1]_{oc} $\;
}
\If{$m = 5$ and F is vanilla AE} { \tcp{add attention}
    sample $l$ from $\{0, \dots, L_{max} \}$\;
    $F'[l]$ $\gets F'[l] \ \cup $ attention layer\;
}
\Return{$F'$}
\caption{Single Model Mutation }
\label{alg:model_mutate}
\end{algorithm}

\begin{algorithm}
\KwIn{$F_1$, $F_2$}
$F_1' \gets copy(F_1)$\;
$F_2' \gets copy(F_2)$\;

sample $m$ from $\{0,1\}$ \tcp*{type of crossover}
\If{$m = 0$ \tcp*{exchange the layers} }{ 
sample $l$ from $\{0, 1, \dots, \Min{(L_{F_1}, L_{F_2}}) \}$ \;
$F_1'[l] $ $\gets$ $F_2[l]$\;
$F_2'[l] $ $\gets$ $F_1[l]$\;
}
\If{$m = 1 \text{and F1 is vanilla AE} $ } { \tcp{exchange the lengths of models}
    \uIf{$L_{F_1} > L_{F_2}$} {
        $F_1'[L_{F_2}-1]_{oc}$ $\gets$ $F_2[L_{F_2}-1]_{oc}$\;
        \For{$k \in [L_{F_2},\dots,L_{F_1}]$} {
            $F_1'[k]$ $\gets$ $F_2[k]$\;
            $F_2'[k]$ $\gets \emptyset$ \;
        }
    }
    \uElse {
        $F_2'[L_{F_1}-1]_{oc}$ $\gets$ $F_1[L_{F_1}-1]_{oc}$\;
        \For{$k \in [L_{F_1},...,L_{F_2}]$} {
            $F_2'[k]$ $\gets$ $F_1[k]$\;
            $F_1'[k]$ $\gets \emptyset\ $ ;
        }
    }
}
\Return{$F_1', F_2'$}
\caption{Models Crossover}
\label{alg:model_cross}
\end{algorithm}


\subsection{Non-gradient fine-tuning}
\label{subsec:fine_tuning}
The practice of adjusting the weights of pre-trained models through gradient-based methods, while widely adopted, can sometimes result in sub-optimal models, as the model may converge to not the best possible solution. A notable contribution in this context is  \cite{nagel2020}, which demonstrates that applying non-gradient fine-tuning after gradient-based optimization can lead to more accurate models. Taking inspiration from this research, we implement a non-gradient fine-tuning process on the best models obtained from the previous steps. We leverage an evolutionary optimization technique, in which genetic operators alter the weights of the models with the objective of enhancing their performance. The mutation operator modifies selected weights by a mutation power $\tau$ with a probability $p_m$. The mutation process can be mathematically represented as follows:
\begin{equation}
    \theta' =  \theta * (1 \pm p_m*\tau),
    \label{eq:fine_tuning_mutation}
\end{equation}
where $\theta$ is a single weight.

\begin{algorithm}
\SetAlgoLined
\SetKwInOut{Input}{inputs}
\SetKw{KwRet}{return}
\SetKw{Return}{return}
\KwResult{$R$ - A set of fine-tuned models}
\KwIn{$F$ - The best model from previous level}
\KwIn{$N_{P}$ - Size of fine-tuning population in every iteration}
\KwIn{$N_{g}$ - Number of iterations in the fine-tuning}
\KwIn{$p_m$ - Mutation probability}
\KwIn{$\tau$ - Mutation power}
$R \gets \emptyset$ \;
$g \gets 0$ \tcp*{Iteration number}
$F_g \gets F$ \;
\While{$g < N_{g}$} {
    $P_g \gets \{ F_{\Theta'_1}, F_{\Theta'_2}, \dots, F_{\Theta'_{N_P}} \}$ where $\theta'$ is created by mutating weights from $F_g$ using eq. \ref{eq:fine_tuning_mutation} with $p_m$ and $\tau$\;
    
    $F_g \gets \displaystyle \argmin_{F_i \in P_g} FP(F_i)$ \;

    \If{$FP(F_g) = 0  \lor \ $\\$
    \ \ \ \forall_{k \in [g-5, g-4, \dots, g-1]} FP(F_k) = FP(F_g)$ \tcp*{is stagnated}}{
        $R \gets R \cup F_g$ \; 
        $F_g \gets \displaystyle \argmax_{F' \in P} \ \  d_F(F, F') $ \tcp*{$d_F(\cdot, \cdot)$ from eq. \ref{eq:distance_fine_tuning}}
     }
    $g \gets g + 1$ \;
}
\Return{$R$}\;
\caption{Weights Fine-tuning \cite{ADNEV}}
 \label{alg:fine}
\end{algorithm}

Algorithm \ref{alg:fine} showcases the fine-tuning process employing a mutation strategy to optimize the model. This involves randomly mutating a specified percentage of the model's weights\footnote{The process of altering weights involves randomly (with 50\% probability) selecting a plus or minus sign for each weight adjustment} (Line 5). The effectiveness of each mutated solution is evaluated based on its fitness, calculated as the number of False Positives (FPs) (Line 6). In this context, a sample is identified as an anomaly (FP) if its reconstruction error exceeds the mean value of all reconstruction errors scaled by a constant factor that determines the allowed deviation from this average.
The algorithm then selects the best-performing solutions from this process for inclusion in the new population (Lines 6-11). The best selected models are a base to generate a new set of models in the next iteration. 

The fine-tuning process of a specific model terminates upon meeting one of two possible stagnation conditions: \textit{i)} if a fitness value of zero FPs is achieved (Line 7); or \textit{ii)} if the model exhibits no further improvement over a designated number of iterations (Line 8). When either of these termination conditions is met, the best-performing model is extracted from the population and saved for the inference phase, where its efficacy is assessed using testing data. 

After extracting the best model, the algorithm selects another model to carry out the fine-tuning process in the following iterations. The selection aims at finding a model with the highest distance from the best model to keep its diversity. The distance is computed as the difference between the weight values of two models:
\begin{equation}
    d_\theta(F_i, F_j) =  \sum_{k=0}^{Layers(F_i)}\sqrt{|\theta_{F_{ik}} - \theta_{F_{jk}}|^2}.
    \label{eq:distance_fine_tuning}
\end{equation}

The process of selecting the most diverse model is a critical component of our fine-tuning strategy, as it allows to explore a broader range of potential solutions and mitigates the risk of the optimization process becoming trapped in local minima.
During the fine-tuning phase, we deliberately exclude the use of a crossover operation, as the implementation of crossover in this context would be the same as setting a very high mutation probability ($p_m$). Such a high value of $p_m$ could lead to significant and abrupt shifts in the model weights within a single iteration. This, in turn, might cause the model weights to drift in suboptimal directions, potentially resulting in a significant decrease in the model's accuracy.



\begin{table*}[ht]
\setlength{\tabcolsep}{18pt}
\caption{Experimental results in terms of point-wise $F_1$ score. * -- models were evaluated using Gauss-D scoring function; N/A -- results are not available. The best result for every dataset is in \textbf{bold}.}
\label{tab:main_results}
\begin{tabular}{lcccclc}
\toprule
                     & \textbf{SWAT} & \textbf{WADI} & \textbf{MSL} & \textbf{SMAP} & \textbf{SMD}                       & \textbf{Mean} \\ \midrule
\textbf{GNN}         & 0.81          & 0.57          & 0.30         & 0.33          &                        0.53            &        0.5       \\
\textbf{USAD}        & 0.79          & 0.43          & 0.48         & 0.49          & 0.41                               &    0.52           \\
\textbf{CNN 1D}      & 0.78          & 0.27          & 0.44         & 0.52          & 0.41                               &       0.48        \\
\textbf{NASA LSTM*}  & 0.13          & 0.20          & 0.55         & 0.59          & 0.39                               &         0.37      \\
\textbf{UAE*}        & 0.58          & 0.47          & 0.54         & 0.58          & 0.55                               &            0.54   \\
\textbf{LSTM-AE}     & 0.45          & 0.33          & 0.54         & 0.53          & 0.53                               &        0.48       \\
\textbf{LSTM-VAE}    & 0.42          & 0.50          & 0.49         & 0.49          & 0.54                               &          0.49     \\
\textbf{TCN AE}      & 0.43          & 0.43          & 0.55         & 0.55          & 0.59                               &         0.51      \\
\textbf{BeatGAN}     & 0.48          & 0.46          & 0.53         & 0.57          & 0.55                               &      0.52         \\
\textbf{OCAN}        & 0.15          & 0.0           & 0.30         & 0.28          & 0.46                               &       0.24        \\
\textbf{DAGMM}       & 0.0           & 0.13          & 0.14         & 0.17          & 0.37                               &       0.16        \\
\textbf{OmniAnomaly} & 0.15          & 0.24          & 0.41         & 0.38          & 0.52                               &        0.34       \\
\midrule
\textbf{AD-NEv}      & 0.82          & 0.62          & 0.57         & 0.77          & 0.56                               &        0.67       \\ \hline
\textbf{AD-NEv++ GNN}      &   \textbf{0.84}          & \textbf{0.59}         &  \textbf{0.32}        &    \textbf{0.35}          &     \textbf{0.52}      &       \textbf{0.56}        \\ 
\textbf{AD-NEv++}    & \textbf{0.84} & \textbf{0.64}    & \textbf{\textbf{0.59}}   & \textbf{0.79}    & \multicolumn{1}{c}{\textbf{0.59}} & \textbf{0.69}    \\ \bottomrule
\end{tabular}
\end{table*}

A summary of our experimental results is shown in Table \ref{tab:main_results}. The table shows the extracted results with different models on five popular benchmark anomaly detection datasets: WADI, SWAT, MSL, SMAP, and SMD.
The last column presents the average point-wise F1-score considering all datasets. The last two rows show the results for AD-NEv++ in two versions. The first one operates exclusively on the graph network architecture (GNN), whereas the second one is a version of AD-NEv++ working for all possible configurations.

It is noteworthy that, after the evolution stage, our approach to optimize graph autoencoder archiectures (AD-NEV++ GNN)\footnote{In our experiments. graph-based autoencoder evolution was performed without subspace evolution to provide a reasonable execution time.} was capable of improving results for all datasets compared to conventional graph autoencoders (GNN). The improvements can be summarized as follows: WADI: from $0.57$ to $0.59$, SWAT: from $0.81$ to $0.84$, MSL: from $0.3$ to $0.32$, SMAP: from $0.33$ to $0.35$, and SMD: from $0.53$ to $0.56$. All these results were achieved with window size range from 3 to 7 and graph layers dimensions from 64 to 512.

Another interesting result emerges when comparing different variants of the framework with different model architectures. 
We observe that after using additional skip and dense connection in the evolution of the models (AD-NEv++), it was possible to improve the results of the basic formulation of the framework (AD-NEv), resulting in the best performance with $4$ datasets. In the case of the SMD dataset, AD-NEv appears as the second-best performing method with a score of 0.56, whereas the TCN AE model has the best result - $0.59$. AD-NEv++ improves the baseline solution by $2\%$ for WADI, SWAT, MSL, and SMAP. In the case of SMD, the results obtained are equal to the best-performing TCN AE model. It is noteworthy that the improvement on the SWAT dataset was achieved with the exploitation of dense connections. We observe that the best performing AD-NEv++ model has the same length as its AD-NEv counterpart, but it contains dense connections between selected encoder and decoder layers (last encoder and first decoder layer). For MSL, SMAP and SMD, the overall performance was also sensibly improved by introducing the attention mechanism. In some machines from MSL, SMAP and SMD datasets, the  models improved AD-NEv results by incorporating attention and fully connected layers (in about 20\% of machines). This result shows that considering the novel types of layers available in AD-NEv++ is extremely valuable in enhancing the model performance on the anomaly detection task. 
The best model maximum encoder length for all datasets was 6. The AD-NEv++ vanilla autoencoder neuroevolution was run on subspaces generated by original AD-NEv framework, \cite{ADNEV}. 

Tables \ref{tab:gnn_time} and \ref{tab:adnev_time} present the execution times for single model training and the whole evolution process for AD-NEv++ GNN and AD-NEv++, respectively. We can observe that the evolution time for AD-NEv++ is much longer than its GNN counterpart (for example, 18h vs 131.8h). This result is expected, since AD-NEv++ explores a wider search space including skip connections, dense connections, and attention. The next aspect, is that graph-based autoencoder training is faster than vanilla autoencoder. However, the evolution times appear still within reasonable limits. Another aspect impacting the evolution time are parameters such as population size and number of iterations, which are presented in Tables \ref{tab:gnn_params} and \ref{tab:adnev_params}. The number of populations used in the experiments ranges from 16 to 30. The number of iterations for which the neuroevolution process was carried out is between 20 and 60. A lower number of iterations was performed for less complex datasets, where the dataset for a single machine is relatively small (SMD, SMAP and MSL). 

Moreover, we can observe that training times depend on the size and the complexity of each dataset, where the most challenging dataset (WADI) requires around five times the execution time than the smallest dataset (SMD).
Finally, it is worth mentioning that the method is still fully scalable as shown in \cite{ADNEV}. By using a larger number of GPUs, the evolution time can be significantly reduced.
All presented results are presented as a mean values from three measurements. The AD-NEv framework was implemented with Python and Py-Torch. All presented calculations were executed leveraging Nvidia Tesla V100-SXM2-32GB GPUs.

\begin{table*}
\centering
\caption{AD-NEv++ GNN execution time: Single model training and overall evolution stage.}
\label{tab:gnn_time}
\begin{tabular}{lcc}
\toprule
                     & \textbf{Train time} & \textbf{Evolution time} \\
                      \midrule
\textbf{WADI}                & {70(min)}         &  {18(h)}   \\ 
\textbf{SWAT}    & {20(min)}    & {7(h)}  \\ 
\textbf{MSL}                & {14(min)}         &  {3(h)}   \\
\textbf{SMAP}    & {40(min)}    & {{6(h)}}   \\ 
\textbf{SMD}    & {14(min)}    & {{3(h)}}   \\ 
\bottomrule
\end{tabular}
\end{table*}

\begin{table}[]
\centering
\caption{AD-NEv++ execution time: Single model training and overall evolution stage.}
\label{tab:adnev_time}
\begin{tabular}{lcc}
\toprule
                     & \textbf{Train time} & \textbf{Evolution time} \\
                      \midrule
\textbf{WADI}                & {1.8(h)}         &  {131.8(h)}   \\ 
\textbf{SWAT}    & {1 (h)}    & {{74(h)}}   \\ 
\textbf{MSL}                & {30(min)}         &  {18.8(h)}   \\ 
\textbf{SMAP}    & {1.5(h)}    & {58.2(h)}   \\ 
\textbf{SMD}    & {28(min)}    & {40(h)}   \\ \bottomrule
\end{tabular}
\end{table}

\begin{table}[]
\centering
\caption{AD-NEv++ GNN: Evolution parameters.}
\label{tab:gnn_params}
\begin{tabular}{lccc}
\toprule
                     & \textbf{population size} & \textbf{iterations} & \textbf{mutation rate}\\
                      \midrule
\textbf{WADI}                & {20}         &  {60}  & 0.02 \\ 
\textbf{SWAT}    & {20}    & {{30}} & 0.02  \\ 
\textbf{MSL}    & {20}    & {{20}} &  0.02 \\  
\textbf{SMAP}    & {20}    & {{20}} &  0.02 \\ 
\textbf{SMD}    & {20}    &  
 {{20}} &  0.02 \\ \bottomrule
\end{tabular}
\end{table}

\begin{table}[]
\centering
\caption{AD-NEv++ evolution parameters.}
\label{tab:adnev_params}
\begin{tabular}{lccc}
                     & \textbf{Population size} & \textbf{Iterations} & \textbf{Mutation rate}\\
                      \midrule
\textbf{WADI}                & {30}         &  {32}  &  0.02 \\ 
\textbf{SWAT}    & {30}    & {{32}} &  0.02 \\ 
\textbf{MSL}    & {20}    & {20} &  0.02\\ 
\textbf{SMAP}    & {20}    & {20} &  0.02\\ 
\textbf{SMD}    & {16}    & {{50}} &  0.02\\ \bottomrule
\end{tabular}
\end{table}

\section{Conclusions and future works}
In this paper, we proposed AD-nEV++, a multi-architecture neuroevolution framework for anomaly detection. 
Our work builds upon existing neuroevolution-based anomaly detection approaches and introduces new capabilities. Our neuroevolution framework performs the simultaneous evolution of models and subspaces, and effectively supports new model architectures such as graph-based autoencoders. Moreover, it provides a new level of abstraction to support optional layers, such as attention, skip connections, and dense connections.   
Our experimental results show that the innovative characteristics of AD-NEv++ allow us to learn and optimize robust autoencoder-based model architectures. We show that AD-NEv++ is capable of outperforming state-of-the-art approaches, including graph-based autoencoders, with popular anomaly detection benchmark datasets.
In future work, we will investigate the evolution of GAN-based models as an additional supported architecture in the framework. Moreover, the framework will be equipped with online learning capabilities, focusing on fast and adaptive model training and adaptation.


\bibliographystyle{unsrt}  
\bibliography{references}

\end{document}